\documentclass[a4paper]{article}

\usepackage{INTERSPEECH_v2}

\title{Advances in Joint CTC-Attention based End-to-End Speech Recognition with a Deep CNN Encoder and RNN-LM}
\name{Takaaki Hori$^1$, Shinji Watanabe$^1$, Yu Zhang$^2$, William Chan$^3$ }

\address{
  $^1$Mitsubishi Electric Research Laboratories\\
  $^2$Massachusetts Institute of Technology\\
  $^3$Carnegie Mellon University}
\email{\{thori,watanabe\}@merl.com, yzhang87@mit.edu, williamchan@cmu.edu}

\begin{document}

\maketitle
\begin{abstract}
  We present a state-of-the-art end-to-end Automatic Speech Recognition (ASR) model.
  We learn to listen and write characters with a joint Connectionist Temporal Classification (CTC) and attention-based encoder-decoder network.
  The encoder is a deep Convolutional Neural Network (CNN) based on the VGG network.
  The CTC network sits on top of the encoder and is jointly trained with the attention-based decoder.
  During the beam search process, we combine the CTC predictions, the attention-based decoder predictions and a separately trained LSTM language model.
  We achieve a 5-10\% error reduction compared to prior systems on spontaneous Japanese and Chinese speech, and our end-to-end model beats out traditional hybrid ASR systems.
 %The joint CTC-attention-based end-to-end automatic speech recognition (ASR) effectively utilizes both advantages of connectionist temporal classification (CTC) and attention-based encoder-decoders,　and performs better than the individual approaches.
 %In our previous work, we designed the joint CTC-attention model as a combination of an encoder network with stacked bidirectional long short-term memory (LSTM) layers, a decoder network with attention-based LSTM layers, and a CTC network that shared the encoder network.  
 %In this paper, we extend our model by incorporating a VGGNet, a very deep convolutional neural network (CNN), into the encoder and a pre-trained large-scale LSTM language model into the decoder.
 %We further improves the beam search process by using CTC prefix probabilities during decoding, while our previous implementation used the CTC probabilities in a rescoring step.
 %We have applied the extended end-to-end system to two ASR benchmarks (spontaneous Japanese and Mandarin Chinese) and obtained 5-10\% error reduction from the prior system. Finally, our system has achieved better performance than state-of-the-art hybrid ASR systems without linguistic resources.
\end{abstract}

\section{Introduction}
\label{sec:intro}
% general background
Automatic Speech Recognition (ASR) is currently a mature set of technologies that have been widely deployed, resulting in great success in interface applications such as voice search \cite{sainath-icassp-2015}.
A typical ASR system is factorized into several modules including acoustic, lexicon, and language models based on a probabilistic noisy channel model \cite{jelinek1976continuous}.
Over the last decade, dramatic improvements in acoustic and language models have been driven by machine learning techniques known as deep learning \cite{hinton2012deep}.  

% conventional hybrid ASR and its problem
However, current systems lean heavily on the scaffolding of complicated legacy architectures that grew up around traditional techniques, including Hidden Markov Model (HMM), Gaussian Mixture Model (GMM), Deep Neural Networks (DNN), followed by sequence discriminative training \cite{Povey_ASRU2011}.
%For example, when we build an acoustic model from scratch, we have to first build hidden Markov model (HMM) and Gaussian mixture model (GMM) followed by deep neural networks (DNN). Furthermore, we often apply state-level minimum Bayes risk (sMBR) training after lattice generation for the entire training data to improve the recognition accuracy.
We also need to build a pronunciation dictionary and a language model, which require linguistic knowledge, and text preprocessing such as tokenization for some languages without explicit word boundaries.
Finally, these modules are integrated into a Weighted Finite-State Transducer (WFST) for efficient decoding.
Consequently, it is quite difficult for non-experts to use/develop ASR systems for new applications, especially for new languages.  

% end-to-end ASR	
End-to-end ASR has the goal of simplifying the above module-based architecture into a single-network architecture within a deep learning framework, in order to address the above issues. End-to-end ASR methods typically rely only on paired acoustic and language data without linguistic knowledge, and train the model with a single algorithm. Therefore, the approach potentially makes it possible to build ASR systems without expert knowledge.
		
There are two major types of end-to-end architectures for ASR:  attention-based methods use an attention mechanism to perform alignment between acoustic frames and recognized symbols \cite{chorowski2014end,chan2015listen,lu2016training,chan-interspeech-2016,chan-iclr-2017}, and Connectionist Temporal Classification (CTC), uses Markov assumptions to efficiently solve sequential problems by dynamic programming \cite{graves2014towards,miao2015eesen,amodei2015deep}.
While CTC requires several conditional independence assumptions to obtain the label sequence probabilities, the attention-based methods do not use those assumptions. This property is advantageous to sequence modeling, but the attention mechanism is too flexible in the sense that it allows extremely non-sequential alignments like the case of machine translation, although the alignments are usually monotonic in speech recognition. 	

To solve this problem, we have proposed joint CTC-attention-based end-to-end ASR \cite{kim2016joint_icassp2017}, which effectively utilizes a CTC objective during training of the attention model. Specifically, we attach the CTC objective to an attention-based encoder network as a regularization technique, which also encourages the alignments to be monotonic. In our previous work, we demonstrated the approach improves the recognition accuracy over the individual use of CTC or attention-based method \cite{kim2016joint_icassp2017}.

In this paper, we extend our prior work by incorporating several novel extensions to the model, and investigate the performance compared to traditional hybrid systems. 
The extensions we introduced are as follows.
\begin{enumerate}
	\item Joint CTC-attention decoding: In our prior work, we used the CTC objective only for training. In this work, we use the CTC probabilities for decoding in combination with the attention-based probabilities. We propose two methods to combine their probabilities, one is a rescoring method and the other is a one-pass method.
	\item Deep Convolutional Neural Network (CNN) encoder: We incorporate a VGG network in the encoder network, which is a deep CNN including 4 convolution and 2 max-pooling layers \cite{simonyan2014very}. 
	\item Recurrent Neural Network Language Model (RNN-LM): We combine an RNN-LM network in parallel with the attention decoder, which can be trained separately or jointly, where the RNN-LM is trained with character sequences.
    %  without word-level knowledge
    % remove this comment because the character sequence implies word sequences which implies word level knowledge
    % alex graves actually ran this exp before and char lms dont suffer much compared to workd lms (personally i disagree tho)
\end{enumerate}
Although the efficacy of a deep CNN encoder has already been demonstrated in end-to-end ASR \cite{zhang2016very,zhang2017towards}, the other two extensions have not been experimented with yet. We present experimental results showing efficacy of each technique, and finally we show that our joint CTC-attention end-to-end ASR achieves performance superior to several state-of-the-art hybrid ASR systems in Spontaneous Japanese and Mandarin Chinese tasks.
% , without using the external linguistic resources.  	
% removing this because its confusing to the reader.

\section{Joint CTC-attention}
\label{sec:joint-ctc-attention}
In this section, we explain the joint CTC-attention framework, which utilizes both benefits of CTC and attention during training \cite{kim2016joint_icassp2017}. 
%--------------------------------------
\subsection{Connectionist Temporal Classification (CTC)}
Connectionist Temporal Classification (CTC) \cite{graves2006connectionist} is a latent
variable model that monotonically maps an input sequence to an output sequence of shorter length.
We assume here that the model outputs $L$-length letter sequence $C = \{ c_l \in \mathcal{U} | l = 1, \cdots, L\}$ with a set of distinct characters $\mathcal{U}$.
CTC introduces framewise letter sequence with an additional "blank" symbol $Z = \{ z_t \in \mathcal{U} \cup \text{blank}| t = 1, \cdots, T\}$.
By using conditional independence assumptions, the posterior distribution $p(C|X)$ is factorized as follows:
\begin{align}
& p(C|X) \approx \underbrace{\sum _{Z} \prod _t p(z_t | z _{t-1}, C) p(z_t|X)} _{\triangleq p _{\text{ctc}}(C|X)} p(C) \label{eq:ctc_final}
\end{align}
As shown in Eq. \eqref{eq:ctc_final}, CTC has three distribution components by the Bayes theorem similar to the conventional hybrid ASR case, i.e., framewise posterior distribution $p(z_t|X)$, transition probability $p(z_t | z _{t-1}, C)$, and letter-based language model $p(C)$.
We also define the CTC objective function $p _{\text{ctc}}(C|X)$ used in the later formulation.

The framewise posterior distribution $p(z_t|X)$ is conditioned on all inputs $X$, and it is quite natural to be modeled by using bidirectional long short-term memory (BLSTM):
\begin{align}
p(z_t|X) & = \text{Softmax}(\text{Lin}(\mathbf{h}_t)) \\
\mathbf{h}_t & = \text{BLSTM}(X) \label{eq:ctc_enc}.
\end{align}
$\text{Softmax}(\cdot)$ is a softmax activation function, and $\text{Lin}(\cdot)$ is a linear layer to convert hidden vector $\mathbf{h}_t$ to a $(|\mathcal{U}| + 1)$ dimensional vector ($+1$ means a $\text{blank}$ symbol introduced in CTC).

Although Eq.~\eqref{eq:ctc_final} has to deal with a summation over all possible $Z$, we can efficiently compute this marginalization by using dynamic programming thanks to the Markov property.
In summary, although CTC and hybrid systems are similar to each other due to conditional independence assumptions, CTC does not require pronunciation dictionaries and omits an HMM/GMM construction step.

%--------------------------------------
\subsection{Attention-based encoder-decoder}
\label{sec:att}
Compared with CTC approaches, the attention-based approach does not make any conditional independence assumptions, and directly estimates the posterior $p(C|X)$ based on the chain rule:
\begin{align}
p(C|X) = \underbrace{\prod _l p(c_l|c_{1}, \cdots, c_{l-1}, X)}_{\triangleq p _{\text{att}}(C|X)},
\label{eq:attention}
\end{align}
where $p _{\text{att}}(C|X)$ is an attention-based objective function.
$p(c_l|c_{1}, \cdots, c_{l-1}, X)$ is obtained by 
\begin{align}
p(c_l &|c_{1}, \cdots, c_{l-1}, X)  = \text{Decoder}(\mathbf{r}_l, \mathbf{q}_{l-1}, c_{l-1}) \label{eq:att_dec} \\
\mathbf{h}_t & = \text{Encoder}(X) \label{eq:att_enc} \\
a_{lt} & = \text{Attention}(\{a_{l-1}\}_t, \mathbf{q}_{l-1}, \mathbf{h}_t) \label{eq:att_att} \\
\mathbf{r}_l & = \sum _t a_{lt} \mathbf{h}_t \label{eq:att_r}.
\end{align}
Eq.~\eqref{eq:att_enc} converts input feature vectors $X$ into a framewise hidden vector $\mathbf{h}_t$ in an encoder network based on BLSTM, i.e., $\text{Encoder}(X) \triangleq \text{BLSTM}(X)$.
$\text{Attention}(\cdot)$ in Eq.~\eqref{eq:att_att} is based on a content-based attention mechanism with convolutional features, as described in \cite{chorowski2015attention}.
$a_{lt}$ is an attention weight, and represents a soft alignment of hidden vector $\mathbf{h}_t$ for each output $c_l$ based on the weighted summation of hidden vectors to form letter-wise hidden vector $\mathbf{r}_l$ in Eq.~\eqref{eq:att_r}.
A decoder network is another recurrent network conditioned on previous output $c_{l-1}$ and hidden vector $\mathbf{q}_{l-1}$, similar to RNNLM, in addition to letter-wise hidden vector $\mathbf{r}_l$.
We use $\text{Decoder}(\cdot) \triangleq \text{Softmax}(\text{Lin}(\text{LSTM}(\cdot)))$.

Attention-based ASR does not explicitly separate each module, but
it implicitly combines acoustic models, lexicon, and language models as encoder, attention, and decoder networks, which can be jointly trained as a single deep neural network.
Compared with CTC, attention-based models make predictions conditioned on all the previous predictions, and thus can learn language. However, the cost of using an explicit alignment without monotonic constraints means the alignment can become impaired.

\subsection{Multi-task learning}
\label{sec:mtl}
In \cite{kim2016joint_icassp2017}, we used the CTC objective function as an auxiliary task to train the attention model encoder within the multi-task learning (MTL) framework. This approach substantially reduced irregular alignments during training and inference, and provided improved performance in several end-to-end ASR tasks.

%\begin{figure}[tb]
%	\includegraphics[width=7.5cm]{joint}
%	\caption{Joint CTC-attention based end-to-end framework: the shared encoder is trained by both CTC and attention model objectives simultaneously. 
%		The shared encoder transforms our input sequence $\{\mathbf{x}_t \cdots \mathbf{x}_T\}$ into high level features $H = \{\mathbf{h}_t \cdots \mathbf{h}_T\}$, and the attention decoder generates the letter sequence $\{c_1 \cdots c_L\}$.}
%	\label{fig:arch}
%\end{figure}
%Figure \ref{fig:arch} illustrates the basic architecture of the framework, where the same BLSTM is shared with CTC and attention decoder networks (that is Eqs.~\eqref{eq:ctc_enc} and \eqref{eq:att_enc}, respectively).
The joint CTC-attention shares the same BLSTM encoder with CTC and attention decoder networks.
Unlike the sole attention model, the forward-backward algorithm of CTC can enforce monotonic alignment between speech and label sequences during training.
%which well regularizes a shared encoder network so that the network training is converged quickly. 
That is, rather than solely depending on the data-driven attention mechanism to estimate the desired alignments in long sequences, the forward-backward algorithm in CTC helps to speed up the process of estimating the desired alignment.
The objective to be maximized is a logarithmic linear combination of the CTC and attention objectives, i.e., $p_\text{ctc} (C|X)$ in Eq.~\eqref{eq:ctc_final} and $p_\text{att} (C|X)$ in Eq.~\eqref{eq:attention}: 
\begin{align}
\label{e12}
\mathcal{L}_\text{MTL} &= \lambda \log p_\text{ctc} (C|X) + (1-\lambda) \log p_\text{att} (C|X),
\end{align}
with a tunable parameter $\lambda: 0 \leq \lambda \leq 1$.

\section{Extended joint CTC-attention}
\label{sec:extended_network}
\begin{figure}[tb]
	\includegraphics[width=7.5cm]{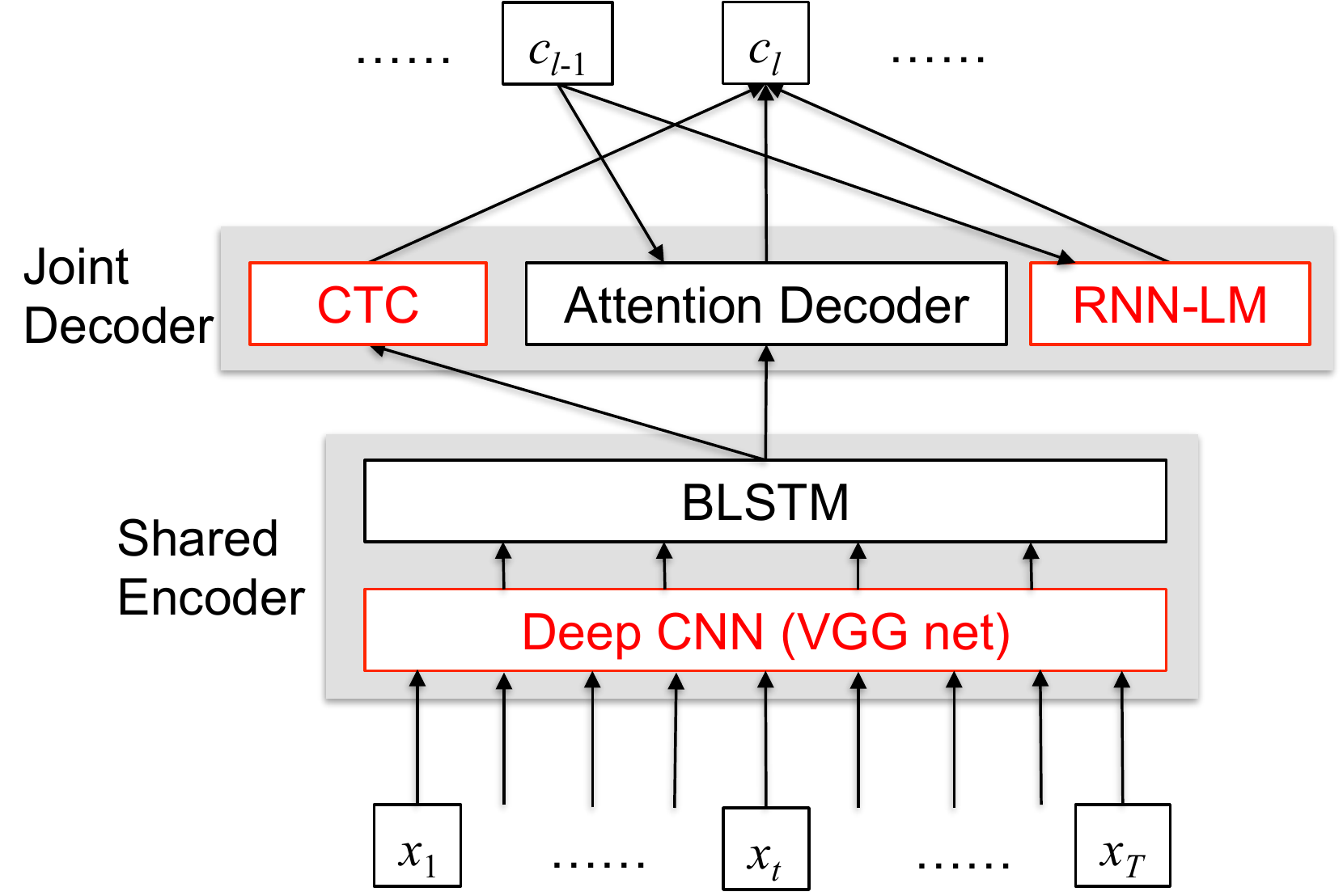}
	\caption{Extended Joint CTC-attention ASR: the shared encoder contains a VGG net followed by BLSTM layers and trained by both CTC and attention model objectives simultaneously. The joint decoder predicts an output label sequence by the CTC, attention decoder and RNN-LM. The extensions made in this paper are colored in red.}
	\label{fig:extended_system}
    \vspace{-0.3cm}
\end{figure}

This section introduces three extensions to our joint CTC-attention end-to-end ASR. Figure \ref{fig:extended_system} shows the extended architecture, which includes joint decoding, a deep CNN encoder and an RNN-LM network.

\subsection{Joint decoding}
\label{sec:decoding}
It is already been shown that the CTC objective helps guide the attention model during training to be more robust and effective, and produce a better model for speech recognition \cite{kim2016joint_icassp2017}. In this section, we propose to use the CTC predictions also in the decoding process.

The inference step of attention-based speech recognition is performed by output-label synchronous decoding with a beam search.
But, we take the CTC probabilities into account to find a better aligned hypothesis to the input speech, i.e. the decoder finds the most probable character sequence $\hat{C}$ given speech input $X$, according to
\begin{align}
\hat{C}=\arg \max_{C\in \mathcal{U}^*} & \left\{\lambda \log p_{\text{ctc}}(C|X)\right. \nonumber \\
& \left. + (1-\lambda)\log p_{\text{att}}(C|X)\right\}.
\label{eq:joint_decoding}
\end{align}

In the beam search process, the decoder computes a score of each partial hypothesis. With the attention model, the score can be computed recursively as
\begin{align}
\alpha_\text{att}(g_l)=\alpha_\text{att}(g_{l-1}) + \log p(c|g_{l-1},X),
\label{eq:score_recursion}
\end{align}
where $g_l$ is a partial hypothesis with length $l$, and $c$ is the last character of $g_l$, which is appended to $g_{l-1}$, i.e. $g_l=g_{l-1}\cdot c$. The score for $g_l$ is obtained as the addition of the original score $\alpha(g_{l-1})$ and the conditional log probability given by the attention decoder in \eqref{eq:att_dec}.
During the beam search, the number of partial hypotheses for each length is limited to a predefined number, called a {\it beam width}, to exclude hypotheses with relatively low scores, which dramatically improves the search efficiency.

However, it is non-trivial to combine CTC and attention-based scores in the beam search, because the attention decoder performs it character-synchronously while CTC does it frame-synchronously.
To incorporate CTC probabilities in the score, we propose two methods. 
One is a rescoring method, in which the decoder first obtains a set of complete hypotheses using the beam search only with the attention model, and rescores each hypothesis using Eq. \eqref{eq:joint_decoding}, where $p_\text{ctc}(C|X)$ can be computed with the CTC forward algorithm.
The other method is a one-pass decoding, in which we compute the probability of each partial hypothesis using CTC and the attention model.
Here, we utilize the CTC prefix probability \cite{graves2008thesis} defined as the cumulative probability of all label sequences that have $g_l$ as their prefix: 
\begin{align}
p(g_l,\dots|X)=\sum_{\nu \in (\mathcal{U}\cup \{\text{\tt <eos>}\})^+} P(g_l \cdot \nu|X),
\end{align}
and we obtain the CTC score as
\begin{align}
\alpha_\text{ctc}(g_l) = \log p(g_l,\dots|X),
\end{align}
where $\nu$ represents all possible label sequences except the empty string, and {\tt <eos>} indicates the end of sentence.
The CTC score can not be obtained recursively as in Eq. \eqref{eq:score_recursion}, but it can be computed efficiently by keeping the forward probabilities over input frames for each partial hypothesis. Then it is combined with $\alpha_\text{att}(g_l)$ using $\lambda$.

\subsection{Encoder with Deep CNN}
\label{sec:deepcnn}
Our encoder network is boosted by using deep CNN, which is motivated by the prior studies \cite{zhang2017towards,zhang2016very}.
We use the initial layers of the VGG net architecture \cite{simonyan2014very} followed by BLSTM layers in the encoder network.
We used the following 6-layer CNN architecture:
\begin{align*}
& \text{Convolution2D}(\text{\# in}=3, \text{\# out}=64, \text{filter}=3 \times 3) \\
& \text{Convolution2D}(\text{\# in}=64, \text{\# out}=64, \text{filter}=3 \times 3) \\
& \text{Maxpool2D}(\text{patch}=3 \times 3, \text{stride}=2 \times 2) \\
& \text{Convolution2D}(\text{\# in}=64, \text{\# out}=128, \text{filter}=3 \times 3) \\
& \text{Convolution2D}(\text{\# in}=128, \text{\# out}=128, \text{filter}=3 \times 3) \\
& \text{Maxpool2D}(\text{patch}=3 \times 3, \text{stride}=2 \times 2)
\end{align*}
The initial three input channels are composed of the spectral features, delta, and delta delta features.
Input speech feature images are downsampled to $(1/4 \times 1/4)$ images along with the time-frequency axises through the two max-pooling (\text{Maxpool2D}) layers.

\subsection{Decoder with RNN-LM}
We combine an RNN-LM network in parallel with the attention decoder, which can be trained separately or jointly, where the RNN-LM is trained with character sequences without word-level knowledge.
Although the attention decoder implicitly includes a language model as in Eq. \eqref{eq:att_dec}, we aim at introducing language model states purely dependent on the output label sequence in the decoder, which potentially brings a complementary effect.

As shown in Fig. \ref{fig:extended_system}, the RNN-LM probabilities are used to predict the output label jointly with the decoder network.
The RNN-LM information is combined at the logits level or pre-softmax. If we use a pre-trained RNN-LM without any joint  training, we need a scaling factor. If we train the model jointly with the other networks, we may combine their pre-activations before the softmax without a scaling factor as this is learnt. In effect, the attention-based decoder learns to use the LM prior.
 
Although it is possible to apply the RNN-LM as a rescoring step, we combine the RNN-LM network in the end-to-end model because we do not wish to have an additional rescoring step. Also, we can view this as a single large neural network model, even if parts of it are separately pretrained. Furthermore, the RNN-LM can be trained jointly with the encoder and decoder networks.
	
%--------------------------------------
\section{Experiments}
%--------------------------------------
We used Japanese and Mandarin Chinese ASR benchmarks to show the effectiveness of the extended joint CTC-attention approaches.
% Going to remove the rationality below... doesn't add much to the paper. This will make the paper look even stronger, and I think alphabet languages should actually be easier because of the phonetic information i alphabets .. its actually more impressive i think to do so well in these languages.	f
%The main reason for choosing these two languages is that those ideogram languages have relatively shorter lengths for letter sequences than those in alphabet languages, which reduces computational complexities greatly, and makes it easy to handle context information in a decoder network.
%Our preliminary investigation shows that Japanese and Mandarin Chinese end-to-end ASR can be easily scaled up, and shows state-of-the-art performance without using various tricks developed in English tasks.

%\subsection{Mandarin telephone speech}
%\label{sec:hkust}
%\subsection{Conditions}
The Japanese task is lecture speech recognition using the Corpus of Spontaneous Japanese (CSJ) \cite{maekawa2000spontaneous}.
CSJ is a standard Japanese ASR task based on a collection of monologue speech data including academic lectures and simulated presentations.
It has a total of 581 hours of training data and three types of evaluation data, where each evaluation task consists of 10 lectures (totally 5 hours).
The Chinese task is HKUST Mandarin Chinese conversational telephone speech recognition (MTS) \cite{liu2006hkust}.
%We demonstrated ASR experiments on HKUST Mandarin Chinese conversational telephone speech recognition (MTS) \cite{liu2006hkust}.
It has 5 hours recording for evaluation, and we extracted 5 hours from training data as a development set, and used the rest (167 hours) as a training set.

As input features, we used 80 mel-scale filterbank coefficients with pitch features as suggested in \cite{ghahremani2014pitch,miao2016empirical} for the BLSTM encoder, and adding their delta and delta delta features for the CNN BLSTM encoder  \cite{zhang2016very}.
The encoder was a 4-layer BLSTM with 320 cells in each layer and direction, and linear projection layer is followed by each BLSTM layer. 
The 2nd and 3rd bottom layers of the encoder read every second hidden state in the network below, reducing the utterance length by the factor of 4 (subsampling). 
When we used the VGG architecture, as described in Section \ref{sec:deepcnn} as the CNN BLSTM encoder, the following BLSTM layers did not subsample the input features.
We used the location-based attention mechanism \cite{chorowski2015attention}, where the 10 centered convolution filters of width 100 were used to extract the convolutional features.
The decoder network was a 1-layer LSTM with 320 cells.
%We also built a RNN-LM as a 1-layer LSTM with 800 cells, which was first trained separately using the transcription, combined with the decoder network, and optionally re-trained with the encoder, decoder and CTC networks jointly.
We also built an RNN-LM as a 1-layer LSTM for each task, where
the CSJ model had 1000 cells and the MTS model had 800 cells.
Each RNN-LM was first trained separately using the transcription, combined with the decoder network, and optionally re-trained with the encoder, decoder and CTC networks jointly. Note that there is no extra text data been used here but we believe more untranscribed data definitely can further improve the results.

The AdaDelta algorithm \cite{zeiler2012adadelta} with gradient clipping \cite{pascanu2012difficulty} was used for the optimization.
We used the $\lambda = 0.1$ for CSJ and the $\lambda = 0.5$ for MTS in training and decoding based on our preliminary investigation.
The beam width was set to 20 in decoding under all conditions.
The joint CTC-attention ASR was implemented by using the Chainer deep learning toolkit \cite{tokui2015chainer}.

%In decoding, we also added a result of the coverage-term based decoding \cite{chorowski2016towards}, as discussed in Section \ref{sec:decoding} ($\eta = 1.5, \tau = 0.5, \gamma =-0.6$ for attention model and $\eta = 1.0, \tau = 0.5, \gamma =-0.1$ for MTL), since it was difficult to eliminate the irregular alignments during decoding by only tuning the maximum and minimum lengths and length penalty (we set the minimum and maximum lengths of output sequences by 0.0 and 0.4 times input sequence lengths, respectively and set $\gamma = 0.6$ in Table \ref{tb:hkust_cer}).
%--------------------------------------
\begin{table}[tb]
	\begin{center}
		%\vspace{0mm}
		\caption{Character Error Rate (CER) for conventional attention and proposed joint CTC-attention end-to-end ASR.
			Corpus of Spontaneous Japanese speech recognition (CSJ) task.}
		\label{tb:csj_cer}
		%\vspace{1mm}
		%\scalebox{0.75}{
		\begin{tabular}{l|c|c|c}
			Model     & Task1 & Task2 & Task3 \\
			\hline
			Attention &  11.4 & 7.9 & 9.0 \\
			MTL       &  10.5 & 7.6 & 8.3 \\
			MTL + joint decoding (rescoring) & 10.1 & 7.1 & 7.8 \\
            MTL + joint decoding (one-pass) &  10.0 & 7.1 & 7.6 \\
			MTL-large + joint dec. (one-pass) & \textbf{8.4} & \textbf{6.2} & \textbf{6.9} \\
			+ RNN-LM (separate) & \textbf{7.9} & \textbf{5.8} & \textbf{6.7} \\
			\hline
%			GMM-discr. \cite{moriya2015kaldi}$^*$ & 11.2         & 9.2         & 12.1         \\
			DNN-hybrid \cite{moriya2015kaldi}$^*$ & 9.0         & 7.2         & 9.6         \\
			DNN-hybrid                                          & 8.4         & 6.9         & 7.1         \\
			CTC-syllable \cite{kanda2016maximum}  & 9.4 & 7.3 & 7.5 \\
			\multicolumn{4}{r}{($^*$using only 236 hours for acoustic model training)}        
		\end{tabular}
	\end{center}
    \vspace{-0.7cm}
\end{table}
%--------------------------------------
\begin{table}[tb]
	\begin{center}
		%\vspace{0mm}
		\caption{Character Error Rate (CER) for conventional attention and proposed joint CTC-attention end-to-end ASR. HKUST Mandarin Chinese conversational telephone speech recognition (MTS) task.}
		\label{tb:hkust_cer}
		%\vspace{1mm}
		%\scalebox{0.75}{
		\begin{tabular}{l|c|c}
			Model & dev & eval \\
			\hline
			Attention     & 40.3 & 37.8 \\
			MTL           & 38.7 & 36.6 \\
			+ joint decoding (rescoring)  & 35.9 & 34.2 \\
			+ joint decoding (one-pass)   & 35.5 & 33.9 \\
			+ RNN-LM (separate)           & 34.8 & 33.3 \\
			+ RNN-LM (joint training)   & \textbf{33.6} & \textbf{32.1} \\
			%\hline
            MTL+joint dec. (speed perturb., one-pass) & 32.1 & 31.4 \\
            + MTL-large & 31.0 & 29.9 \\
            + RNN-LM (separate) & 30.2 & 29.2 \\
            MTL+joint dec. (speed perturb., one-pass) & - & -  \\
			+ VGG net             & 30.0  & 28.9  \\
			+ RNN-LM (separate) & \textbf{29.1} & \textbf{28.0} \\
			\hline
			DNN-hybrid  & -- & 35.9 \\
			LSTM-hybrid (speed perturb.)  & -- & 33.5 \\
			CTC with language model \cite{miao2016empirical} & -- & 34.8 \\
			\parbox[c]{5cm}{TDNN-hybrid, lattice-free MMI  (speed purturb.) \cite{povey2016purely}} & -- & 28.2 \\
		\end{tabular}
	\end{center}
    \vspace{-0.6cm}
\end{table}

%\subsection{Results}
Tables \ref{tb:csj_cer} and \ref{tb:hkust_cer} show character error rates (CERs) of evaluated methods in CSJ and MTS tasks, respectively. 
In both tasks, we can see the effectiveness of joint decoding over the baseline attention model and our prior work with multi-task learning (MTL), 
especially showing the significant improvement of the joint decoding with the one-pass method and RNN-LM integration.
We performed retraining of the entire network including the RNN-LM only in MTS task, because of time limitation. The joint training further improved the performance, which reached 32.1\% CER as shown in Table \ref{tb:hkust_cer}.
%Table \ref{tb:hkust_cer} shows the effectiveness of joint decoding over our prior work with multi-task learning (MTL), especially showing the significant improvement of the joint decoding with the one-pass method and RNN-LM integration with joint training, which reaches 32.1\% CER.

We also built a larger network (MTL-large) for CSJ, which had a 6-layer encoder network and an RNN-LM, to compare our method with the conventional state-of-the-art techniques obtained by using linguistic resources.
The state-of-the-art CERs of DNN-sMBR hybrid systems are obtained from the Kaldi recipe \cite{moriya2015kaldi} and a system based on syllable-based CTC with MAP decoding \cite{kanda2016maximum}.
The Kaldi recipe systems originally only use academic lectures (236h) for AM training, but we extended to use all training data (581h).
The LMs were trained with all training-data transcriptions.
Finally, our extended joint CTC-attention end-to-end ASR achieved lower CERs than already reported CERs obtained by the hybrid approaches for CSJ.

In MTS task, we generated more training data by linearly scaling the audio lengths by factors of 0.9 and 1.1 (speed perturb.).
The final model including the VGG net and RNN-LM achieved $\textbf{28.0}$\% without using linguistic resources, which defeats state-of-the-art systems including recently-proposed lattice-free MMI methods. Although we could not apply jointly-trained RNN-LM when using speed perturbation because of time limitation, we hopefully obtain further improvement by joint training.
\section{Conclusion}
%-----------------------------------------------------------------
%In this paper, we presented a state-of-the-art end-to-end ASR model. Our model learns to listen and write characters with a joint Connectionist Temporal Classification (CTC) and attention-based encoder-decoder network.
%  The encoder is a deep Convolutional Neural Network (CNN) based on the VGG network.
%  The CTC network sits on top of the encoder and is jointly trained with the attention-based decoder.
%  During the beam search process, we combine the CTC predictions, the attention-based decoder predictions and a separately or jointly trained LSTM language model.
%Even with these extensions, our model does not require linguistic resources, such as morphological analyzer and pronunciation dictionary, which are essential components of conventional Mandarin Chinese and Japanese ASR systems.
%Our method achieves state-of-the-art performance even when compared to conventional hybrid-HMM systems for Mandarin Chinese and Japanese tasks.

In this paper, we proposed a novel approach for joint CTC-attention decoding and RNN-LM integraton for end-to-end ASR model. We also explored deep CNN encoder to further improve the extracted acoustic features. Together, we significantly improved current best end-to-end ASR system without any linguistic resources such as morphological analyzer and pronunciation dictionary, which are essential components of conventional Mandarin Chinese and Japanese ASR systems. Our end-to-end joint CTC-attention model outperforms hybrid systems without the use of any explicit language model on our Japanese task. Moreover, our method achieves state-of-the-art performance when combined with a pretrained character level language model on both Chinese and Japanese, even when compared to conventional hybrid-HMM systems. We note that despite using a pretrained RNN-LM, the model can be seen as one big neural network with a seperately pretrained components. Finally, we emphasize the text data we used to train our RNN-LM is from the same text data in the labelled audio data, we did not use any extra text. We believe our model can be further improved using vast quantities of unlabelled data to pretrain a RNN-LM and subsequently jointly trained with our model.

	\bibliographystyle{IEEEtran}
	
	\bibliography{refs}
	
	% \begin{thebibliography}{9}
	% \bibitem[1]{Davis80-COP}
	%   S.\ B.\ Davis and P.\ Mermelstein,
	%   ``Comparison of parametric representation for monosyllabic word recognition in continuously spoken sentences,''
	%   \textit{IEEE Transactions on Acoustics, Speech and Signal Processing}, vol.~28, no.~4, pp.~357--366, 1980.
	% \bibitem[2]{Rabiner89-ATO}
	%   L.\ R.\ Rabiner,
	%   ``A tutorial on hidden Markov models and selected applications in speech recognition,''
	%   \textit{Proceedings of the IEEE}, vol.~77, no.~2, pp.~257-286, 1989.
	% \bibitem[3]{Hastie09-TEO}
	%   T.\ Hastie, R.\ Tibshirani, and J.\ Friedman,
	%   \textit{The Elements of Statistical Learning -- Data Mining, Inference, and Prediction}.
	%   New York: Springer, 2009.
	% \bibitem[4]{YourName17-XXX}
	%   F.\ Lastname1, F.\ Lastname2, and F.\ Lastname3,
	%   ``Title of your INTERSPEECH 2017 publication,''
	%   in \textit{Interspeech 2017 -- 18\textsuperscript{th} Annual Conference of the International Speech Communication Association, August 20?24, Stockholm, Sweden, Proceedings, Proceedings}, 2017, pp.~100--104.
	% \end{thebibliography}
	
\end{document}